\documentclass[sigconf]{acmart-me}

\usepackage{booktabs}
\usepackage{url}
\usepackage{color}
\usepackage{enumitem}
\setcopyright{rightsretained}

\acmDOI{}

\acmISBN{}

\acmConference[MediaEval'21]{Multimedia Evaluation Workshop}{December 13-15 2021}{Online}
\acmYear{2021}
\copyrightyear{}

\acmPrice{}

\begin{document}
\title{Sports Video: Fine-Grained Action Detection and Classification of Table Tennis Strokes from Videos for MediaEval 2021}

\author{Pierre-Etienne Martin\textsuperscript{1}, Jordan Calandre\textsuperscript{2}, Boris Mansencal\textsuperscript{3}, Jenny Benois-Pineau\textsuperscript{3},\\
Renaud P\'eteri\textsuperscript{2}, Laurent Mascarilla\textsuperscript{2}, Julien Morlier\textsuperscript{4}}

\affiliation{\textsuperscript{1}CCP Department, Max Planck Institute for Evolutionary Anthropology, D-04103 Leipzig, Germany\\
\textsuperscript{2}MIA, La Rochelle University, La Rochelle, France\\
\textsuperscript{3}Univ. Bordeaux, CNRS,  Bordeaux INP, LaBRI, Talence, France\\
\textsuperscript{4}IMS, University of Bordeaux, Talence, France}

\email{mediaeval.sport.task@diff.u-bordeaux.fr}

\renewcommand{\shortauthors}{P-e Martin et al.}
\renewcommand{\shorttitle}{Sports Video Task}

\begin{abstract}
Sports video analysis is a prevalent research topic due to the variety of application areas, ranging from multimedia intelligent devices with user-tailored digests up to analysis of athletes' performance. The Sports Video task is part of the MediaEval 2021 benchmark. This task tackles fine-grained action detection and classification from videos. The focus is on recordings of table tennis games. 
Running since 2019, the task has offered a classification challenge from untrimmed video recorded in natural conditions with known temporal boundaries for each stroke. This year, the dataset is extended and offers, in addition, a detection challenge from untrimmed videos without annotations. This work aims at creating tools for sports coaches and players in order to analyze sports performance. Movement analysis and player profiling may be built upon such technology to enrich the training experience of athletes and improve their performance.
\end{abstract}

\maketitle

\section{Introduction}
\label{sec:intro}

Action detection and classification are one of the main challenges in computer vision~\cite{PeChapSpringer:2021}. Over the last few years, the number of datasets and their complexity dedicated to action classification has drastically increased~\cite{PeThesis2020}.
Sports video analysis is one branch of computer vision and applications in this area range from multimedia intelligent devices with user-tailored digests, up to analysis of athletes' performance~\cite{LenhartL:2018, TT:TTNet:2020, Sport:EvaluationMotion:2017}. A large amount of work is devoted to the analysis of sports gestures using motion capture systems. However, body-worn sensors and markers could disturb the natural behavior of sports players. This issue motivates the development of methods for game analysis using non-invasive equipment such as video recordings from cameras.

 The Sports Video Classification project was initiated by the Sports Faculty (STAPS) and the computer science laboratory LaBRI of the University of Bordeaux, and the MIA laboratory of La Rochelle University\footnote{This work was supported by the New Aquitania Region through CRISP project - ComputeR vIsion for Sports Performance and the MIRES federation.}.
 This project aims to develop artificial intelligence and multimedia indexing methods for the recognition of table tennis activities. The ultimate goal is to evaluate the performance of athletes, with a particular focus on students, to develop optimal training strategies. To that aim, the video corpus named \texttt{TTStroke-21} was recorded with volunteer players.

Datasets such as UCF-101~\cite{Dataset:UCF101:2012}, HMDB~\cite{Dataset:HMDB:2011, Dataset:JHMDB:2013}, AVA~\cite{Dataset:AVA:2018} and Kinetics~\cite{Dataset:Kinetics:2017,Dataset:Kinetics600:2018,Dataset:Kinetics700:2019,Dataset:Kinetics700:2020,Dataset:AVA_Kinetics:2020} are being use in the scope of action recognition with, year after year, an increasing number of considered videos and classes. Few datasets focus on fine-grained classification in sports such as FineGym~\cite{Dataset:Gym:2020} and TTStroke21~\cite{PeMTAP:2020}.

To tackle the increasing complexity of the datasets, we have on one hand methods getting the most of the temporal information: for example, in~\cite{LiuH:2019}, where spatio-temporal dependencies are learned from the video using only RGB data. And on the other hand, methods combining other modalities extracted from videos, such as the optical flow~\cite{NN:I3DCarreira:2017, NN:Laptev:2018, PeICIP:2019}. The inter-similarity of actions - strokes - in \texttt{TTStroke-21} makes the classification task challenging, and both cited aspects shall be used to improve performance.

The following sections present the Sport task this year and its specific terms of use. Complementary information on the task may be found on the dedicated page from the MediaEval website\footnote{\url{https://multimediaeval.github.io/editions/2021/tasks/sportsvideo/}}.

\section{Task description}
\label{sec:task}

This task uses the \texttt{TTStroke-21} database~\cite{PeMTAP:2020}. This dataset is constituted of recordings of table tennis players performing in natural conditions. This task offers researchers an opportunity to test their fine-grained classification methods for detecting and classifying strokes in table tennis videos. Compared to the Sports Video 2020 edition, this year, we extend the task with detection, and enrich the data set with new and more diverse stroke samples. The task now offers two subtasks. Each subtask has its own split of the dataset, leading to different train, validation, and test sets.
\par
Participants can choose to participate in only one or both subtasks and submit up to five runs for each. The participants must provide one XML file per video file present in the test set for each run. The content of the XML file varies according to the subtask. Runs may be submitted as an archive (zip file), with each run in a different directory for each subtask. Participants should also submit a working notes paper, which describes their method and indicates if any external data, such as other datasets or pretrained networks, was used to compute their runs. The task is considered fully automatic: once the videos are provided to the system, results should be produced without human intervention. Participants are encouraged to release their code publicly with their submission. This year, a baseline for both subtasks was shared publicly~\cite{mediaeval/Martin21/baseline}.

\subsection{Subtask 1 - Stroke Detection}

Participants must build a system that detects whether a stroke has been performed, whatever its class, and extract its temporal boundaries. The aim is to distinguish between moments of interest in a game (players performing strokes) from irrelevant moments (time between strokes, picking up the ball, having a break…). This subtask can be a preliminary step for later recognizing a stroke that has been performed.
\par
Participants have to segment regions where a stroke is performed in the provided videos. Provided XML files contain the stroke temporal boundaries (frame index of the videos) related to the train and validation sets. We invite the participants to fill an XML file for each test video in which each stroke should be temporally segmented frame-wise following the same structure.
\par
For this subtask, the videos are not shared across train, validation, and test sets; however, a same player may appear in the different sets. The Intersection over Union (IoU) and Average Precision (AP) metrics will be used for evaluation. Both are usually used for image segmentation but are adapted for this task:
\begin{itemize}
\item \textbf{Global IoU:} the frame-wise overlap between the ground truth and the predicted strokes across all the videos.

\item \textbf{Instance AP:} each stroke represents an instance to be detected. Detection is considered True when the IoU between prediction and ground truth is above an IoU threshold. $20$ thresholds from $0.5$ to $0.95$ with a step of $0.05$ are considered, similarly to the COCO challenge~\cite{CocoChallenge2014}. This metric will be used for the final ranking of participants.
\end{itemize}

\subsection{Subtask 2 - Stroke Classification}

This subtask is similar to the main task of the previous edition~\cite{mediaeval/Martin20/task}. This year the dataset is extended, and a validation set is provided.
\par
Participants are required to build a classification system that automatically labels video segments according to a performed stroke. There are 20 possible stroke classes. The temporal boundaries of each stroke are supplied in the XML files accompanying each video in each set. The XML files dedicated to the train and validation sets contain the stroke class as a label, while in the test set, the label is set to ``\texttt{Unknown}''. Hence for each XML file in the test set, the participants are invited to replace the default label ``\texttt{Unknown}'' by the stroke class that the participant's system has assigned according to the given taxonomy.
\par
For this subtask, the videos are shared across the sets following a random distribution of all the strokes with the proportions of 60\%, 20\% and 20\% respectively for the train, validation and test sets.
All submissions will be evaluated in terms of global accuracy for ranking and detailed with per-class accuracy.
\par
Last year, the best global accuracy (31.4\%) was obtained by~\cite{DBLP:conf/mediaeval/Nguyen-TruongCN20} using Channel-Separated CNN. \cite{DBLP:conf/mediaeval/MartinBMPM20} is second (26.6\%) using 3D attention mechanism and \cite{DBLP:conf/mediaeval/SatoA20} third (16.7\%) using pose information and cascade labelling method. Improvement has been observed compared to the previous edition~\cite{PeMETask:2019} with a best accuracy of 22.9\%~\cite{PeMEWork:2019}. This improvement seems to be correlated by various factors such as: i) multi-modal methods, ii) deeper and more complex CNN capturing simultaneously spatial and temporal features, and iii) class decision following a cascade method.

\section{Dataset description}
\label{sec:dataset}

\begin{figure}
    \includegraphics[width=.32\linewidth]{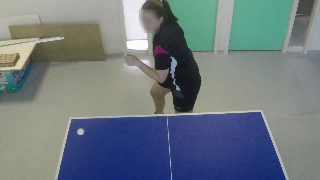}
    \includegraphics[width=.32\linewidth]{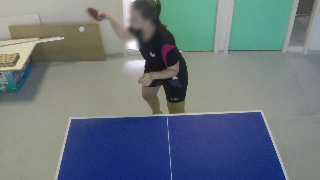}
    \includegraphics[width=.32\linewidth]{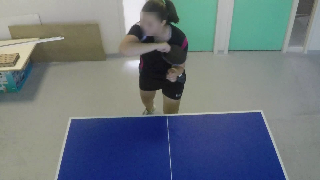}\\
    \vspace{-5pt}
    \caption{Key frames of a same stroke from \texttt{TTStroke-21}}
    \label{fig:dataset}
    \vspace{-15pt}
\end{figure}

The dataset has been recorded at the STAPS using lightweight equipment. It is constituted of player-centered videos recorded in natural conditions without markers or sensors, see Fig~\ref{fig:dataset}. Professional table tennis teachers designed a dedicated taxonomy. The dataset comprises 20 table tennis stroke classes: height services, six offensive strokes, and six defensive strokes.
The strokes may be divided in two super-classes: \texttt{Forehand} and \texttt{Backhand}.
\par
All videos are recorded in MPEG-4 format. We blurred fhe faces of the players for each original video frame using OpenCV deep learning face detector, based on the Single Shot Detector (SSD) framework with a ResNet base network. A tracking method has been implemented to decrease the false positive rate. The detected faces are blurred, and the video is re-encoded in MPEG-4.
\par
Compared with Sports Video 2020 edition, this year, the data set is enriched with new and more diverse video samples. A total of 100 minutes of table tennis games across 28 videos recorded at 120 frames per second is considered. It represents more than $718$~$000$ frames in HD ($1920 \times 1080$). 
An additional validation set is also provided for better comparison across participants. This set may be used for training when submitting the test set's results. Twenty-two videos are used for the Stroke Classification subtask, representing $1017$ strokes randomly distributed in the different sets following the previously given proportions. The same videos are used in the train and validation sets of the Segmentation subtask, and six additional videos, without annotations, are dedicated to its test set.
\vspace{-5pt}

\section{Specific terms of use}
\label{sec:conditions}

Although faces are automatically blurred to preserve anonymity, some faces are misdetected, and thus some players remain identifiable. In order to respect the personal data of the players, this dataset is subject to a usage agreement, referred to as {\it Special Conditions}.
The complete acceptance of these {\it Special Conditions} is a mandatory prerequisite for the provision of the Images as part of the MediaEval 2021 evaluation campaign. A complete reading of these conditions is necessary and requires the user, for example, to obscure the faces (blurring, black banner) in the video before use in any publication and to destroy the data by October 1st, 2022.
\vspace{-5pt}

\section{Discussions}
\label{sec:discussion}
This year the Sports Video task of MediaEval proposes two subtasks: i)~Detection and ii)~Classification of strokes from videos. Even if the players' faces are blurred, the provided videos still fall under particular usage conditions that the participants need to accept. Participants are encouraged to share their difficulties and their results even if they seem not sufficiently good. All the investigations, even when not successful, may inspire future methods.

\begin{acks}
Many thanks to the players, coaches, and annotators who contributed to \texttt{TTStroke-21}.
\end{acks}

\bibliographystyle{ACM-Reference-Format}
\bibliography{sigproc} 

\end{document}